\documentclass[conference]{IEEEtran}
\IEEEoverridecommandlockouts
\usepackage{amsmath,amssymb,amsfonts}
\usepackage{ragged2e}
\usepackage{algpseudocode}
\usepackage{algorithm}
\usepackage{graphicx}
\usepackage{textcomp}
\usepackage{xcolor}
\def\BibTeX{{\rm B\kern-.05em{\sc i\kern-.025em b}\kern-.08em
    T\kern-.1667em\lower.7ex\hbox{E}\kern-.125emX}}
\usepackage[sorting=nty]{biblatex}
\usepackage{multirow}

\IEEEoverridecommandlockouts\IEEEpubid{\makebox[\columnwidth]{978-1-6654-8045-1/22/\$31.00~\copyright~2022 IEEE \hfill} \hspace{\columnsep}\makebox[\columnwidth]{ }}

\begin{document}

\title{An adaptive human-in-the-loop approach to emission detection of Additive Manufacturing processes and active learning with computer vision\\
{\footnotesize \textsuperscript{}}
\thanks{}
}

\author{\IEEEauthorblockN{Xiao Liu}
\IEEEauthorblockA{\textit{I-Form} \\
\textit{School of Computing} \\
\textit{Dublin City University}\\
Dublin, Ireland \\
xiao.liu5@mail.dcu.ie}
\and
\IEEEauthorblockN{Alan F. Smeaton}
\IEEEauthorblockA{\textit{School of Computing} \\
\textit{Dublin City University}\\
Dublin, Ireland \\
alan.smeaton@dcu.ie}
\and
\IEEEauthorblockN{Alessandra Mileo}
\IEEEauthorblockA{\textit{School of Computing} \\
\textit{Dublin City University}\\
Dublin, Ireland \\
alessandra.mileo@dcu.ie}
\and

}

\maketitle

\begin{abstract}
Recent developments in in-situ monitoring and process control in Additive Manufacturing (AM), also known as 3D-printing, allows the collection of large amounts of emission data during the build process of the parts being manufactured. This data can be used as input into 3D and 2D representations of the 3D-printed parts. However the analysis and use, as well as the characterization of this data still remains a manual process. The aim of this paper is to propose an adaptive human-in-the-loop approach using Machine Learning techniques that automatically inspect and annotate the emissions data generated during the AM process. More specifically, this paper will look at two scenarios: firstly, using convolutional neural networks (CNNs) to automatically inspect and classify  emission data collected by  in-situ monitoring and secondly, applying Active Learning techniques to the developed classification model to construct a human-in-the-loop mechanism in order to accelerate the labeling process of the emission data.
The CNN-based approach relies on transfer learning and fine-tuning, which makes the approach applicable to other industrial image patterns.
The adaptive nature of the approach is enabled by uncertainty sampling strategy to automatic selection of samples to be presented to human experts for annotation. 

\end{abstract}


\section{Introduction}
Due to the rapid development in the Additive Manufacturing industry and its associated monitoring technology, as well as the improved availability of data storage, the size of the data generated, collected and stored during the manufacturing process has increased dramatically. As a result, it is common that  raw data collected by the monitoring system is not labeled and the large amount unlabeled data  usually exceeds the capability of manually analyzing and labeling. This creates an impediment to exploiting this data fully. For example it is  challenging to train high performance supervised deep learning models with insufficient properly labelled data samples as is the case here. 

Transfer learning is commonly used in computer vision  tasks with deep learning such as classification on text \cite{dai2007transferring} and images \cite{shaha2018transfer}, video recognition \cite{su2014transfer}, hand gesture recognition \cite{cote2017transfer} and human action \cite{sargano2017human} recognition.   
Transfer learning is a method that first performs training and modifying the weights in the model using data from a source domain and later applies the trained model to a target domain that is different from the source. This is done to allow rapid progress in re-training  the model in the target domain and support improved performance of related tasks such as text and or classification \cite{Pan2010ASO}.

Active learning \cite{settles.tr09} is an approach based on the idea that a machine learning approach trained on a reduced amount of labelled data can provide more accurate results if it can select what data to be trained on. When active learning is used, the system asks queries to a human annotator for interactive labelling of new data points to improve the performance of the trained  model. An approach based on active learning prioritises the most informative data to submit as queries to the human annotator for labeling. Such data points are automatically selected and they potentially have the highest impact on the supervised training of the machine learning model in accelerating the training process.

In this paper, both transfer learning and active learning are key components of the design methodology, and the combination of transfer learning and active learning allows leveraging small amounts of labelled data to improve the performance of the training process of the selected deep learning model. 

\section{Literature review}
\subsection{Review of transfer learning and fine tuning in deep learning}

To develop approaches to automatic inspection during the AM process that are both good on adaptability and high in accuracy, several methods  based on Convolutional Neural Networks (CNNs) have been proposed. \cite{soukup2014convolutional} and \cite{weimer2016design} have each applied CNNs in their work and achieved higher classification accuracy than conventional machine learning algorithms. However, a major problem with that prior work is that to train a deep CNN from scratch will require a large amount of data for training. This is one of the characteristics of deep learning, the need for a lot of training data \cite{patel2020upsurge}.  Thus, one of the challenges in this work is to find a solution that allows us to apply deep CNNs when only a limited amount of labelled data is available for training.

In \cite{zeiler2014visualizing}, \cite{oquab2014learning} and \cite{hafemann2015transfer} the authors have used transfer learning to address the problem of little training data availability by using pre-trained weights from a source network to set the weights of a target network and then using the target network to fulfil the task of feature extraction. However, the performance improvements with these approaches to using transfer learning depend on the fact that there is similarity between the source and target domains in their tasks. The authors of \cite{yosinski2014transferable} have pointed out that if there is a significant difference between the source and target domains, then such  transfer learning with fixed transferred weights can yield less accurate results, and this is the case in the work reported to date during the AM inspection process. To further address the problem in the performance of  transfer learning when using dissimilar target and source data, the authors of \cite{kim2017transfer} proposed a method that applies fine-tuning on a VGG 16 network \cite{simonyan2014very}. The results shows significant improvements on  overall classification performance.

\subsection{Review of active learning}

Although applying machine learning in image-based inspection of AM processes can be a powerful approach to high accuracy defect detection, a major challenge is to create sufficiently large labelled datasets for the training process. Manually creating  large training datasets is time consuming, expensive, and often infeasible in industrial production settings. Thus, it is important to have an alternative  approach to address this problem. For this, active learning can be considered to allow us to start with a limited amount of labelled training data and to enlarge the labeled dataset based on the learning outcome in previous steps.

\begin{figure}[htbp]
\centerline{\includegraphics[width=0.5\textwidth]{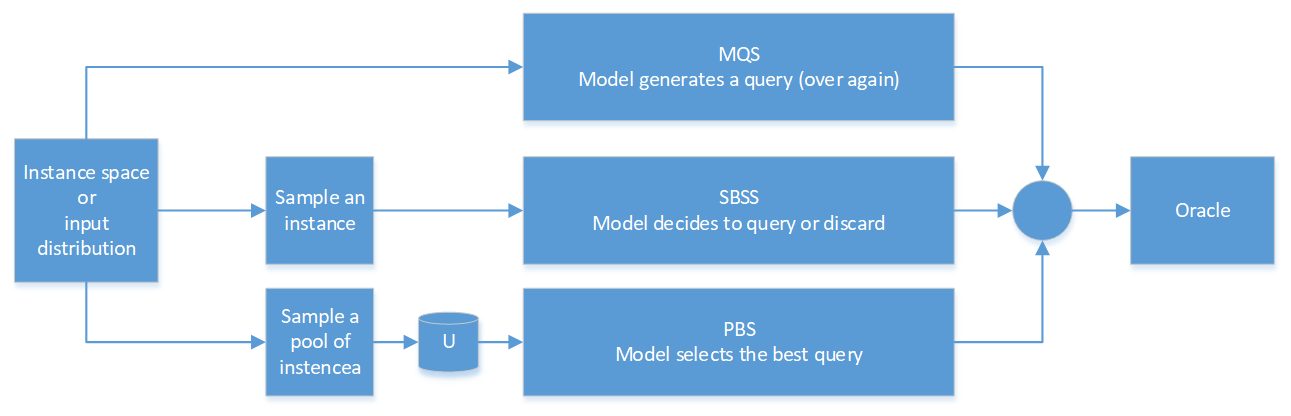}}
\caption{The three scenarios in active learning: MQS, SBSS and PBS}
\label{fig:figS_AL}
\end{figure}

There are three main scenarios for active learning in which the labels of instances can be queried by the learner. These are Membership Query Synthesis (MQS), Stream-Based Selective Sampling (SBSS) and Pool-Based Sampling (PBS). In the different scenarios, unlabeled instances are queried to be labeled by the oracle which is normally referred to the human annotator. The three scenarios can be summarised as shown in Figure~\ref{fig:figS_AL}.

In Membership Query Synthesis (MQS), the learner generates instances that are similar to the instances of current learning target following certain underlying natural distributions.  Generated instances are then sent to the oracle as a query to label. This scenario has a limitation that the queried instances generated by the learner may not be suitable for a human annotator. In the research of Lang and Baum \cite{angluin1988queries}, they employed MQS with human oracles to train a neural network for the classification of handwritten characters. However, many of the query images of characters generated by the learner are not recognizable by the human and cannot be labelled properly.

Stream-Based Selective Sampling (SBSS) makes the assumption that getting an unlabelled instance is free. Based on this assumption, we then select each unlabelled instance, one at a time. According to the informativeness of the selected instance, the learner will determine whether the label of the instance should be queried. Normally query strategies are used to determine the informativeness of the instances.

Pool-Based Sampling (PBS) assumes that there is a large pool of unlabelled data, according to some informativeness measure, also known as Query Strategies, from which instances are drawn from the pool to be queried. The informativeness measure is applied to all instances in the pool to select the most informative instances of which labels will be requested. PBS is the most common scenario in active learning. The main difference between SBSS and PBS is that SBSS makes query decisions individually by scanning through the data sequentially While PBS will first analysis the all instances in the pool and then selecting the best query. In the next section, several common query strategies used to evaluate the best queries are illustrated. 

\subsection{Uncertainty sampling query strategies}
The main difference between active learning and passive learning is the action of the query. Thus it is very important to have proper strategies that can be used to measure the informativeness of the unlabelled instances in order to create the best query from which to identify further instances tobe annotated manually. As stated in \cite{settles.tr09} uncertainty sampling is the most commonly used query strategy. In this section, three methodologies that are used in uncertainty sampling query strategies are illustrated.

The first method to be introduced is called the least confidence query strategy. Least confidence takes the highest probability for the prediction of each data point, then sorts them from smaller to larger. The formal expression to prioritise using least confidence is show as:

$$
x^*_{LC} = {\underset{x}{\operatorname{argmax}}}\Bigl(  1 - P_\theta(\widehat{y}|x) \Bigr)\\
$$
\noindent
where:

$$
\widehat{y} = {\underset{y}{\operatorname{argmax}}}\Bigl( P_\theta(y|x)\Bigr) 
$$

\noindent 
Margin sampling, as the second method, considers the difference between the first and the second highest probability. The data points with the lower margin sampling score would be the ones labelled  first; these are the data points the model is least certain about between the most probably and the next-to-most probable class. Formally, the expression of Margin sampling is:
$$
x^*_{M} = {\underset{x}{\operatorname{argmin}}}\Bigl(P_\theta(\widehat{y}_1|x) - P_\theta(\widehat{y}_2|x)\Bigr)  \\
$$

\noindent 
where:

$$
\widehat{y}_1 and \widehat{y}_2 
$$
are the first and second most probable classes.

\noindent

\
Entropy is a concept that originates in thermodynamics. This concept can be reused to measure the certainty of a model. If a model is highly certain about a class for a given data point, it will probably have a high certainty for a particular class, whereas all the other classes will have low probability. In the case of high entropy it would mean that the model distributes equally the probabilities for all classes as it is not certain at all which class that data point belongs to. It is therefore straightforward to prioritise data points with higher entropy to the ones with lower entropy. Formally, the expression of the entropy score prioritisation is:
$$
x^*_{H} = {\underset{x}{\operatorname{argmax}}}\Bigl( -  \underset{i}{\Sigma}P_\theta(y_i|x)\log P_\theta(y_i|x)\Bigr) \\
$$

\section{Creation of datasets}

\begin{figure}[htb]
    \centering
    \includegraphics[width=0.45\textwidth]{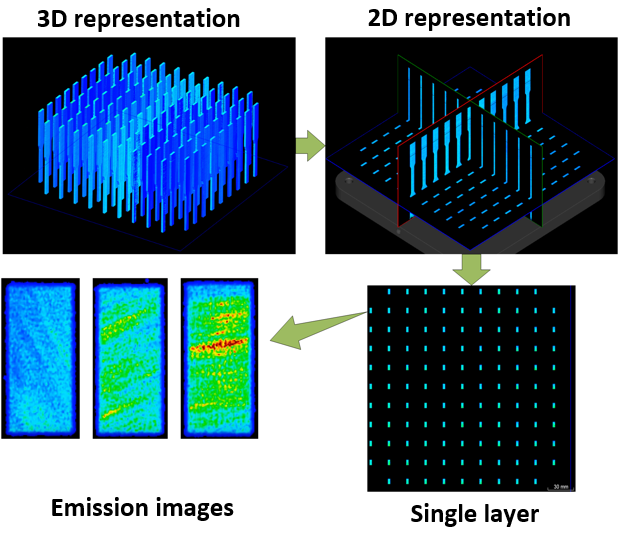}
    \caption{3D to 2D representations of emissions during the AM manufacture process}
    \label{fig:fig2}
\end{figure}

This section describes the creation of the dataset that is used in all the  experiments in this paper.  The data streams collected from the modules which are used in the AM process can be used to build 2D and 3D representations of some of the characteristics or features of the objects being manufactured. Figure~\ref{fig:fig2} shows examples of images formed by analysis of the emissions from the AM process, shown in  2D and 3D part representations.

Originally, the emissions data collected by the sensors in the in-situ monitoring system are formed into 3D points clouds shown as the 3D representation of the emissions features of the parts being built. This 3D representation can be further examined as a series of 2D representations, which is detailed as vertical and/or horizontal cutting surfaces to the level of single layers which have been generated during the printing process. These are in the form of 2D images of melt pool emissions. By further zooming in on the 2D images of a single layer, the emissions image of each part for the current layer can be obtained individually and can be used to create an image dataset for further investigation and research into possible defects in the final object

The first set of data used in this research was collected from the 2D representations of the emission images generated by the in-situ monitoring suite described above. The images represent the melt pool conditions of the printed layers in 11 dog-bone shaped testing parts of Titanium alloy (Ti6Al4V) during the AM process. In each part, 1,000 images of layers were selected yielding a total of 11,000 images  manually exported from the in-situ monitoring suit.

Initially, all the raw image data are unlabelled and thus these raw data are not suitable to be used directly for training. For initial inspection, the 11,000 sample images were manually inspected in order to select 150 images as defected samples~\cite{LIU2022b} and another 150 images as normal samples~\cite{LIU2022n}. These were labeled and used to create a dataset considered as the ground truth in the tests. The size of this labelled dataset is relatively small for the training of the ML models. We acknowledge that  automatic methods could have been used to select the most representative sample or investigate the impact of sample selection in the training process. However, in this investigation, we aim at demonstrating the feasibility of the approach with limited training data and we postpone the analysis of sample selection to future work.

\section{Applying active learning to  emission images}
This section is used to illustrate the methodology that applies active learning to the outcome of the deep learning model that was developed based on a VGG-16 architecture using transfer learning with fine-tuning techniques. Note that human experts are included in the learning process via the \textit{queries}, which require experts to explicitly annotate the (automatically selected) most uncertain samples according to the CNN classification output.

\subsection{The CNN based classifier}
The CNN based deep learning model is an important part of our overall learning architecture. It is the core model to build the initial classifier and to be further built into the active learner.

\begin{figure}[htb]
    \centering
    \includegraphics[width=0.45\textwidth]{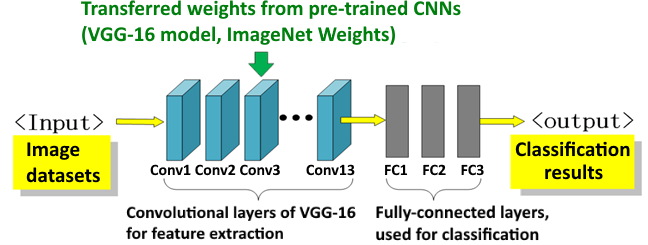}
    \caption{The architecture of the model based on VGG 16 and fully connected layers.}
    \label{fig:fig4_3}
\end{figure}

This approach to create a classifier relies on a transfer learning method in which the 13 convolutional layers from the pre-trained convolutional layers of a VGG 16 model are used for feature extraction and the weights in these layers are transferred weights trained using ImageNet data \cite{deng2009imagenet}. Fine-tuning is subsequently applied to the model to further improve the overall classification accuracy. As illustrated in Figure~\ref{fig:fig4_3}, after the convolutional layers, 2 dense layers with ReLU activation function are added  followed by 1 dense layer as the output layer using Sigmoid as the activation function, since detecting either normal or defect individually for each pattern is a binary classification task.

\begin{figure}[ht]
    \centering
    \includegraphics[width=0.45\textwidth]{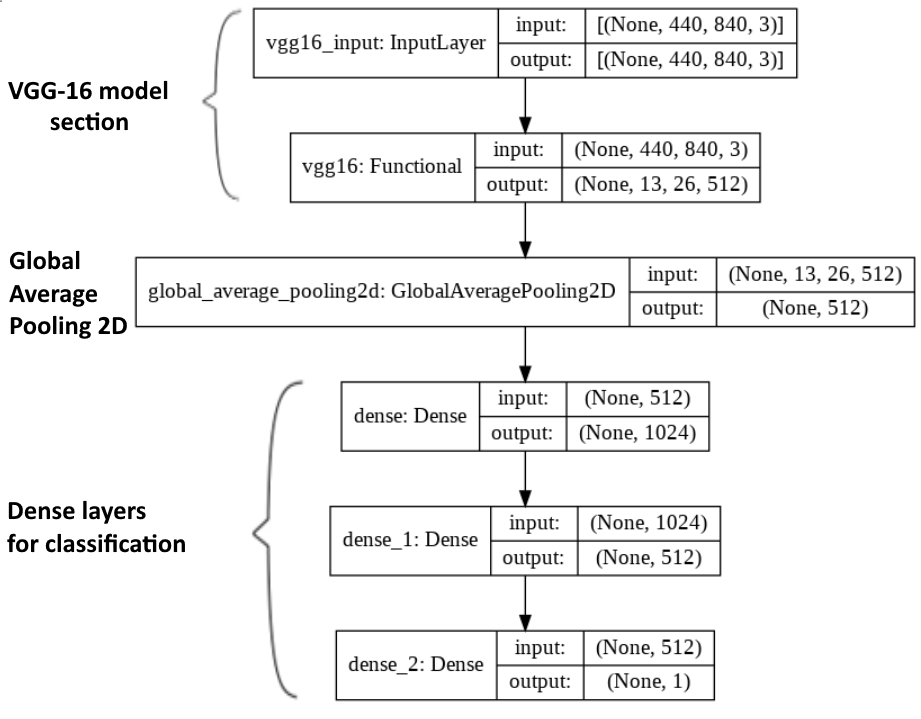}
    \caption{The overall model including the VGG 16 architecture layers, the GAP2D layer and the dense layers for classification. The value “None” in the shape of tensors indicates the dimension is a variable}
    \label{fig:fig4_9}
\end{figure}

To further reduce the spatial dimension of the output of the feature extraction model based on VGG 16, a Global Average Pooling 2D (GAP2D) layer is added between the outcome of the  VGG 16 model used for feature extraction and the dense layers of the classifier. The GAP2D layer receives the output tenser from the VGG 16 model and applies a global average pooling operation for spatial data. By doing this the number of total channels in the output of the GAP2D layer is reduced to 512 and makes it ready to be processed in the dense layers. This is shown in Figure~\ref{fig:fig4_9} which is the overall architecture of the developed deep learning model for the classification tasks.  

\begin{figure}[htb]
    \centering
    \includegraphics[width=0.45\textwidth]{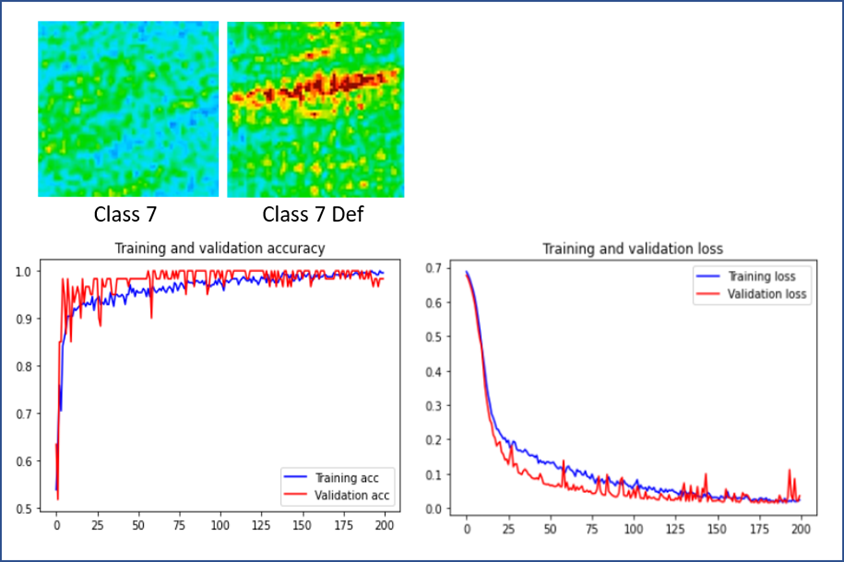}
    \caption{Accuracy and loss for testing on the emission image dataset, using the model with pooling and fine tuning on the last 3 convolutional layers.}
    \label{fig:fig4_8}
\end{figure}

\subsection{The active learning mechanism}
The main model is based on the outcome of the CNN classifier mentioned above, which has been implemented using Keras-tensorflow deep learning Python packages and used as the Learner. To build the whole active-learning mechanism, another ML python package named Scikit-learn was used in combination with Keras. A pseudo-code representation of this active learning mechanism is given as Algorithm~\ref{alg:AL}. In the procedure, there are variables that can be tested to investigate the related impacts on the performance of the overall learning process of the active learning model, which are the initial data for training $X_{init}$ and the number of Queries Q$_{i}$ and the number of queried samples $N$ in each Query. 

\begin{algorithm}
\caption{Active learning with Deep learning model for emission images} \label{alg:AL}
\begin{algorithmic}[1]
\Procedure{AL for emission images }{$ $}
    \State Given (X$_{train}$, y$_{trian}$) = (X$_{l}$,y$_{l}$)
    \State Assemble initial data for training (X$_{init}$, y$_{init}$); (X$_{init}$, y$_{init}$) $\in$ (X$_{train}$, y$_{trian}$)
    \State Create the pool by removing the initial data from the training dataset $X_{pool}$ $\gets$ $X_{train}$ - $X_{init}$
    \State Current dataset used to train the initial classifier $X_{teach}$ = X$_{init}$
    \State Train initial classifier C$_{init}$ using $X_{teach}$
    \For {all queries Q$_{i}$}

        \State Query N samples to assemble $X_{query}$ from $X_{pool}$
        \State Enlarge the dataset for training X$_{teach}$ $\gets$ $X_{teach}$ + $X_{query}$  
        \State Teach the classifier using newly obtained $X_{teach}$ 
        \State Remove queried instances from pool $X_{pool}$ $\gets$ $X_{pool}$ - $X_{query}$
    \EndFor
\EndProcedure
\end{algorithmic}
\end{algorithm}

\subsection{experiment A : Learning with different numbers of samples and queries}
\label{subsect:a}

To start the first experiment, as the initial learner, the CNN based classifier with the transferred weight from ImageNet as mentioned in the previous section was trained using 100 prepared samples for  initial fine-tuning. 100 samples were then removed from the pool as they had already been used in the training and fine-tuning dataset. To efficiently use  computational power, the training length is set as 25 epochs for the reason that according to Figure~\ref{fig:fig4_8} the performance of the classification model shows the most significant improvement before the training of epoch 25. For the same reason, the training length of each query iteration of the active learning process is also set to 25 epochs. 

\begin{figure}[htb]
    \centering
    \includegraphics[width=0.45\textwidth]{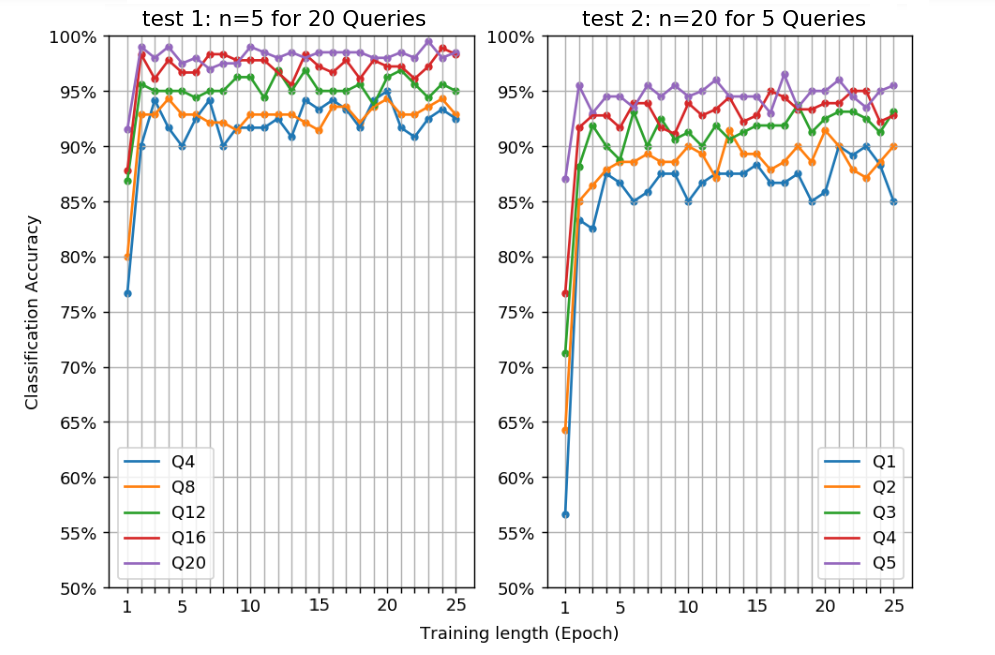}
    \caption{Accuracy of testing on the emission image dataset, using the model with pooling and fine tuning}
    \label{fig:ini100}
\end{figure}

This test aims to investigate the impact of the number of queried samples in each query when the same number of total training samples is available. The query strategy used in this test is the Least Confidence query strategy. The  test was performed with 2 different combinations: in test 1, 5 samples in each query for 20 queries and in test 2, 20 samples in each query for 5 queries. In this way a total of 100 samples are involved in the query/learning process. The result of the classification accuracy during the active learning process of the 2 different setups are shown in  Figure~\ref{fig:ini100}.

\begin{figure}[htb]
    \centering
    \includegraphics[width=0.45\textwidth]{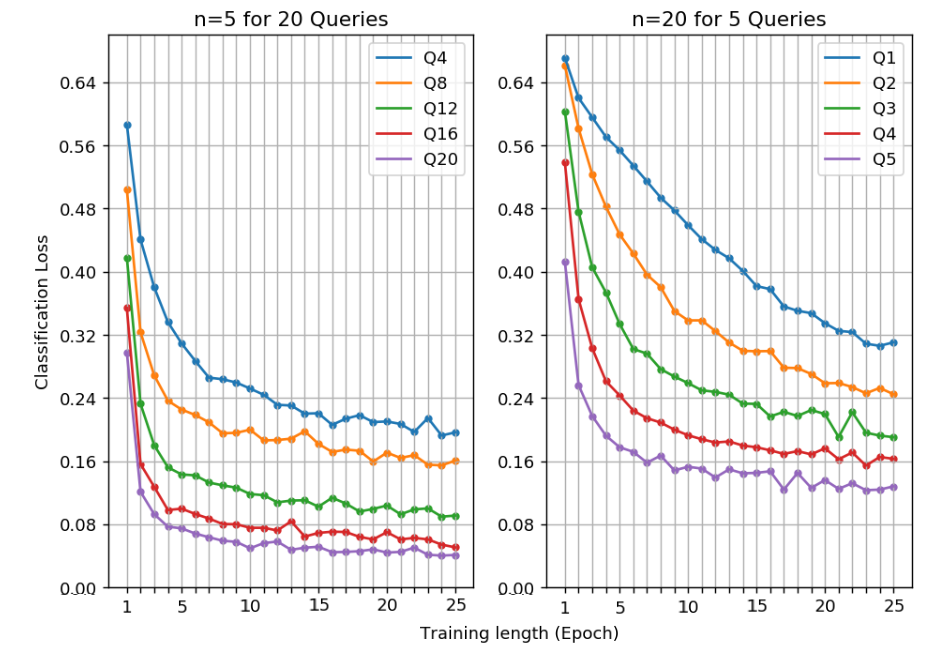}
    \caption{Loss of testing on the emission image dataset, using the model with pooling and fine tuning}
    \label{fig:ini100L}
\end{figure}

For comparison, in the left chart of Figure~\ref{fig:ini100} the lines of the queries Q4 Q8 Q12 Q16 Q20, which are  5 queries out of the total of 20 queries, are selectively shown in order to match the condition of the same total number of samples used in the corresponding 5 queries in the right chart, which are labeled as Q1 to Q5. The loss during the active learning process of the 2 different setups is shown with a similar selection in  Figure~\ref{fig:ini100L}. 

The results show that test1 is generally higher in accuracy when reaching the relatively steady state in each query iteration. When a total of 100 more samples are used in the both tests, test1 shows around 97\% and with less fluctuation in the overall training accuracy while the value in test2  fluctuates at around 95\%.

For the accuracy at first epoch in each query iteration, test1 also shows higher values that starts at over 75\% and reaches about 92\%. In comparison, test2 starts as low as  55\% for Q1 and reaches about 87\%. It is predictable that the accuracy value of Q4 in test1 is considerably higher that Q1 in test2 for the reason that the classifier in test1 is more experienced as it has received training in the previous queries while in test2 the training will have just begun. However, when further  examing more training epochs, when the accuracy no longer improves with more epochs in the current query iteration, where the learner reaches the limit of the learning progress with the same total amount of training samples for example  20 additional samples both in test1 and test2, test1 still shows a higher level of accuracy in general.

From the view of the loss in the active learning process, test1 with fewer samples in each query for more queries than the corresponding setup of test2 results in earlier convergence at a lower value of loss, which means higher confidence in each classification. This  also agrees with the fact that there are less fluctuations in the accuracy at later stages of test1. This behavior is similar to human learning activity: one can compete a learning task with a shorter time when the learning content is less and gain a better understanding after certain rounds of reviewing, until final understanding. There will be further illustration of this in the discussion section.

\subsection{experiment B :The impact of initial training with different numbers of samples} 
\begin{figure}[htb]
    \centering
    \includegraphics[width=0.5\textwidth]{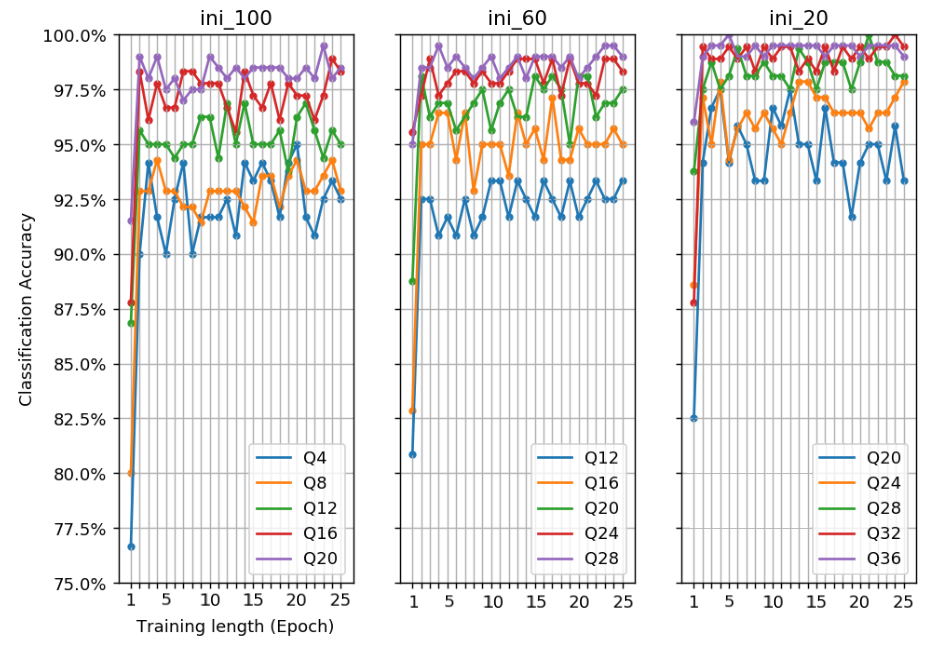}
    \caption{Classification accuracy of the very last query in the learning process with different initial sample sizes}
    \label{fig:iniComAcc}
\end{figure}

In this experiment, we investigate the impact on the overall performance of the learner when changing the number of samples used for the initial training and the number of samples used in the active learning. Three sets of tests are reported where the number of samples for the initial training were set as 20, 60 and 100 respectively while keeping the condition the same as in the last experiment that 5 samples were asked in each query until a total of 200 samples were processed for all  three tests. The results for each test at different total numbers of samples used in the queries, which are 120, 140, 160, 180 and 200 corresponding to line colors blue, yellow, green, red and purple with relevant query number labelled, in Figure~\ref{fig:iniComAcc}.   

\begin{figure}[htb]
    \centering
    \includegraphics[width=0.45\textwidth]{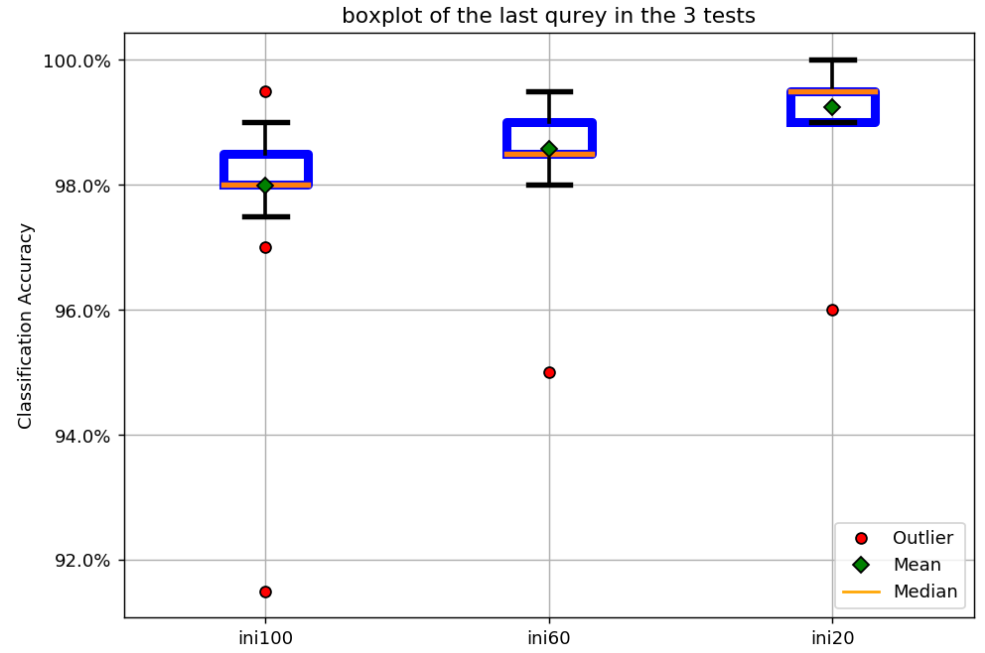}
    \caption{Boxplots of the very last query in the learning process with different initial samples for each of the three tests}
    \label{fig:Box}
\end{figure}

Overall, the results in Figure~\ref{fig:iniComAcc}  show that a greater number of active-selected samples at an early stage of the training can further improve the final accuracy in the training with relatively lower fluctuation in the classifications results. The boxplot of the very last query in the learning process for each of the three tests are shown in Figure~\ref{fig:Box}, where the outliers are reducing in numbers as well as the distances from the outliers to the caps of the corresponding box are reducing while the active selected samples is increasing.     

The mean values for accuracy and the values of standard deviation for each of the three tests are shown in Table~\ref{tab:AandS}. The values were calculated after removing the outliers.

\begin{table}[ht]
\centering
\caption{Average accuracy and standard deviations  in the very last query iterations for each of the three tests.}
\label{tab:AandS}
\begin{tabular}{|c|ccc|}
\hline
\multirow{2}{*}{} & \multicolumn{3}{c|}{Number of initial samples}                \\ \cline{2-4} 
                  & \multicolumn{1}{c|}{100} & \multicolumn{1}{c|}{60} & 20 \\ \hline
\begin{tabular}[c]{@{}c@{}}Average \\ accuracy\end{tabular}   & \multicolumn{1}{c|}{98.25\%} & \multicolumn{1}{c|}{98.73\%} & 99.38\% \\ \hline
\begin{tabular}[c]{@{}c@{}}Standard \\ deviation\end{tabular} & \multicolumn{1}{c|}{0.0055}  & \multicolumn{1}{c|}{0.0045}  & 0.0026  \\ \hline
\end{tabular}
\end{table}

According to Table \ref{tab:AandS}, when a greater proportion of training samples are active-selected in the tests using 20 initial samples (thus, 80 further samples are actively learned for the first 100 training samples), compared to the test with 100 initial random samples, although the improvement in average accuracy may not be significant as around 1\% which is from 98.25\% to 99.38\%, the standard deviation is smaller showing that  overall classification accuracy is more stable if active learning is applied earlier.

\section{discussion}

In real world applications using the active learning process for AM  emission images, the queries need to be confirmed by a human annotator. The workload for the human annotator is thus highly depend on the volume of the total samples queried. 

In experiment A, where we used learning with different numbers of samples and queries as described in Section~\ref{subsect:a} by maintaining the number of the total actively queried samples, the workload for the oracle remains the same. The results from  experiment A with different test setups showed that the method using fewer samples in each query with more queries performed better in the classification tasks. In fact, according to Figure~\ref{fig:ini100}, test 1 out performed test 2 by reaching about 95\% in accuracy using an additional 60 queried samples compared to the 100 queried samples used in test 2. This means that the human annotators are able to obtain the similar performance for classification with 40 less samples identified and thus the approach can  reduce the human workload in the whole active learning process. 

On the other hand, when considering the first epoch in each query iteration, thus usually has a relatively low value compared to the average classification accuracy value of the current loop. This is probably due to  over-fitting caused by the relatively small size of the training dataset, but as the loss decreases, the value of accuracy at the first epoch is approaching to the mean value as  the training dataset enlarges with the iterations. 

For further investigation, the result of experiment B shows that applying active learning at the early stage of the training will improve the learning of the deep model and result in higher final accuracy and confidence of the classification tasks. In future development, when this training approach a high enough  level of  performance, the classification and training processes can be fully automated. In that case, the active learning mechanism can be used to label new data without the help of human oracle and can use the newly obtained data to reinforce the training dataset.  

\section{Conclusions}

In this paper we have designed and tested a deep learning model based on CNNs to automatically identify defects in the AM process of titanium alloy Ti6A14V. In our approach, we leverage the active learning technique to further improve the performance of the training of the classification model while trying not to  increase the potential human workload required in the process. 

Our experiments have demonstrated that our model with transfer learning and active learning can obtain good performances even with relatively low computational power and limited training data and this approach can be further developed to aid in automatic labelling of emission image data. We believe the ability to generate more labelled data using the outcome of our model is necessary, as it not only enables faster convergence (with limited number of epochs), but it also represents a valuable resource to be used by other researchers. For this reason, we are working towards finalising the implementation of an end-to-end framework based on the combination of deep learning with active learning to produce good quality labelled data in other type of AM processes. Such a pipeline can be used for different classification tasks where the availability of labelled dataset is scarce. Currently we are looking into two  such tasks: automatic segmentation of individual melt pool areas and characterisation of porosity structure in the AM processes.

\subsection*{Acknowledgment}

This research is supported by a research grant from Science Foundation Ireland (SFI) under Grant Number 16/RC/3872 and is co-funded under the European Regional Development Fund and by I-Form industry partners. We also wish to thank Prof. Denis Dowling, Director of the Advanced Manufacturing Research Centre (I-Form) and Darragh Egan (I-Form PhD student, University College Dublin) for the provision this data as collected from Renishaw's in-situ monitoring system.

\printbibliography

\end{document}